\documentclass[lettersize,journal]{IEEEtran}
\usepackage{amsmath,amsfonts}
\usepackage{algorithmic}
\usepackage{algorithm}
\usepackage{array}
\usepackage[caption=false,font=footnotesize,labelfont=rm,textfont=rm]{subfig}
\usepackage{textcomp}
\usepackage{stfloats}
\usepackage{url}
\usepackage{verbatim}
\usepackage{graphicx}
\usepackage{cite}
\usepackage{epstopdf}
\usepackage{multirow}
\usepackage{xcolor}
\usepackage{bm}
\usepackage{tabularx}
\usepackage[colorlinks,
linkcolor=blue,
anchorcolor=blue,
urlcolor = blue,		
citecolor=blue]{hyperref}
\usepackage{booktabs}  
\usepackage{colortbl}  
\usepackage{pifont}
\usepackage{threeparttable}

\makeatother

\begin{document}

\title{A Central Motor System Inspired Pre-training Reinforcement Learning for Robotic Control}

\author{Pei Zhang, Zhaobo Hua, Jinliang Ding,~\IEEEmembership{Senior Member,~IEEE}
\thanks{
This work has been submitted to the IEEE for possible publication. Copyright may be transferred without notice, after which this version may no longer be accessible.
	
This work was supported in part by the National Natural Science Foundation of China under Grant 61988101, the National Key R\&D Plan Project under Grant 2022YFB3304700, the 111 Project 2.0 under Grant B08015,  and the Liaoning Province Central Leading Local Science and Technology Development Special Project under Grant 2022JH6/100100055. (Corresponding author: Jinliang Ding)}
\thanks{P. Zhang and Z. Hua, and J. Ding are with the State Key Laboratory of Synthetical Automation for Process Industries, Northeastern University, Shenyang 110819, China (email: pei.zhang088@outlook.com; Bobby\_1752635630@outlook.com;    jlding@mail.neu.edu.cn). 
	
Parts of the figure were drawn by using pictures from Servier Medical Art. Servier Medical Art by Servier is licensed under a Creative Commons Attribution 3.0 Unported License (https://creativecommons.org/licenses/by/3.0/).}}

\markboth{Submission to IEEE Transactions on Systems, Man and Cybernetics: Systems}%
{Shell \MakeLowercase{\textit{et al.}}: A Sample Article Using IEEEtran.cls for IEEE Journals}


\maketitle

\begin{abstract}
The development of intelligent robots requires control policies that can handle dynamic environments and evolving tasks. Pre-training reinforcement learning has emerged as an effective approach to address these demands by enabling robots to acquire reusable motor skills. However, they often rely on large datasets or expert-designed goal spaces, limiting adaptability. Additionally, these methods need help to generate dynamic and diverse skills in high-dimensional state spaces, reducing their effectiveness for downstream tasks.
In this paper, we propose CMS-PRL, a pre-training reinforcement learning method inspired by the Central Motor System (CMS). First, we introduce a fusion reward mechanism that combines the basic motor reward with mutual information reward, promoting the discovery of dynamic skills during pre-training without reliance on external data. Second, we design a skill encoding method inspired by the motor program of the basal ganglia, providing rich and continuous skill instructions during pre-training. Finally, we propose a skill activity function to regulate motor skill activity, enabling the generation of skills with different activity levels, thereby enhancing the robot's flexibility in downstream tasks.
We evaluate the model on four types of robots in a challenging set of sparse-reward tasks. Experimental results demonstrate that CMS-PRL generates diverse, reusable motor skills to solve various downstream tasks and outperforms baseline methods, particularly in high-degree-of-freedom robots and complex tasks.
\end{abstract}

\begin{IEEEkeywords}
  Central motor system, pre-training, reinforcement learning (RL), robotic motor control.
\end{IEEEkeywords}

\section{Introduction}
\IEEEPARstart{W}{ith} the rapid development of technology, the demand for intelligent robots in industries such as manufacturing, services, and healthcare continues to grow\cite{9863891},\cite{10623850}. A core challenge is designing flexible and adaptive control policies to handle dynamic environments and evolving tasks\cite{suomalainen2022survey},\cite{biggs2003survey}. In recent years, reinforcement learning has emerged as a key method for enabling robots to learn flexible behaviors\cite{edmonds2019tale},\cite{9283177}, but it often depends on specific tasks, requiring retraining when tasks or control objects change.

Pre-training reinforcement learning, inspired by human learning modes, has been proposed to reduce training costs for new tasks. It first trains a low-level controller to learn reusable skills and then a high-level controller to manage downstream tasks\cite{peng_ase_2022},\cite{DBLP:conf/iclr/MerelAPTLTHW19},\cite{DBLP:conf/iclr/MerelHGAPWTH19}. 
The main challenge in this process is acquiring a rich set of reusable skills during pre-training.

A direct and effective approach to generating motor skills is to mimic animal behavior. This data-driven approach involves collecting extensive animal motion data, enabling the controller to track a specified posture sequence and thus learn realistic motor skills\cite{DLS},\cite{dnsRL},\cite{peng_ase_2022},\cite{PASP}. However, a given dataset is typically suited to only specific types of robots, and the diversity of skills depends heavily on the quality of the reference data. The goal space method transforms skill learning into a multi-objective tracking problem to eliminate reliance on external data\cite{gehring_hierarchical_2021},\cite{ETG},\cite{HGG}. By designing appropriate goal spaces, this approach enables robots to learn a variety of motion techniques quickly. While this method eliminates data dependence, designing goal spaces requires significant expert experience and testing, particularly when dealing with different types of robots, as goal spaces often need to be tailored specifically.

Unsupervised reinforcement learning offers a solution that does not rely on explicit task goals or reference data\cite{Liu2021BehaviorFT},\cite{park_lipschitz-constrained_2022},\cite{pmlr-v139-liu21b}. This method encourages robots to autonomously collect experience from past motion trajectories and learn skills using intrinsic rewards, making it applicable to a wide range of robots. However, in high-dimensional state spaces, unsupervised learning often results in static skills that are not useful for downstream tasks\cite{eysenbach_diversity_2018},\cite{achiam2017variational}. Some approaches incorporate explicit or implicit prior knowledge to encourage the generation of more dynamic skills\cite{park_lipschitz-constrained_2022},\cite{DBLP:conf/iclr/SharmaGLKH20} (e.g., making large state variations). Still, these methods have primarily proven effective for navigation tasks and may not generalize to other domains.

The limitations of existing methods have prompted researchers to explore the neural mechanisms underlying human motor control\cite{DBLP:conf/iclr/MerelHGAPWTH19},\cite{yuan2023hierarchical}.Human motor flexibility and versatility arise from the Central Motor System (CMS)\cite{bib2}, where the cerebellum and basal ganglia are crucial for motor skill learning, control, coordination, and storage. The cerebellum maintains coordinated movement and generates basic motor abilities, serving as a key center for skill learning and storage\cite{ebner2008cerebellum},\cite{bursztyn2006neural}.
Through synaptic connections, the basal ganglia provide motor program instructions to the cerebellum \cite{bostan2018basal} and regulate voluntary movement by modulating activity levels via internal pathways\cite{cui2013concurrent},\cite{surmeier2013go}.

Building on these insights, we propose a pre-training reinforcement learning method (CMS-PRL) inspired by the CMS mechanisms. This method addresses the challenges of heavy data or expert experience dependence, weak skill dynamics in high-dimensional state spaces, and limited applicability to downstream task types present in existing pre-training reinforcement learning methods. We introduce a fusion reward for pre-training, combining the cerebellum's motor coordination and skill storage capabilities. This reward consists of two components: the basic motor reward and mutual information reward, which facilitate the discovery of dynamic and diverse skills and their storage in a mid-level motor primitive module. We also design a skill encoding method inspired by the basal ganglia's motor program, providing varied random skill instructions for pre-training. Furthermore, we abstract the basal ganglia's regulation of motor activity into a skill activity function (activity level refers to the speed of movement), which we use as a weighting factor for the basic motor reward. This factor enables the separation of motor skills at different levels of activity, allowing robots to demonstrate greater flexibility in handling diverse downstream tasks. Following pre-training, a high-level controller is trained to combine the motor primitive module to address various downstream tasks.

The main contributions of this article are as follows:
\begin{enumerate}
	\item CMS-PRL, a pre-training reinforcement learning method inspired by the intrinsic mechanisms of the CMS. This method can be applied to various types of robots, enabling them to generate rich, reusable skills and efficiently solve a wide range of downstream tasks.
	\item An intrinsic fusion reward that allows robots of differing complexities to discover dynamic and diverse motor skills during pre-training, without relying on expert knowledge or external datasets.
	\item A skill encoding method designed to provide continuous skill vectors during pre-training, enhancing the performance of robots in complex environments.
	\item A skill activity function that enables robots to generate motor skills with varying levels of activity, improving their adaptability in different types of downstream tasks.
\end{enumerate}

We conduct an in-depth experimental analysis of the proposed method using four different types of robots across a challenging set of sparse-reward tasks. The results demonstrate that CMS-PRL generates rich dynamic motor skills during pre-training and outperforms baselines across different tasks. Notably, it shows greater adaptability in high-degree-of-freedom robots and complex task environments.

\section{Related Work}
\label{sec2}
\textit{Data-Driven Skill Learning}:
Learning motor skills from data has been proven to be an effective method. Park et al.\cite{PASP} proposed a network-based algorithm for learning control policies from minimally-labeled human motion data, and Peng et al.\cite{DLS} introduced a two-level hierarchical control framework to learn environment-aware locomotor skills with minimal prior knowledge. Peng et al.\cite{peng_ase_2022} later extended this to the ASE framework, which embeds a large number of skills into the controller from unstructured data. However, these methods depend on the quality and specificity of datasets, limiting skill applicability to particular robots (e.g., quadruped data is only useful for quadruped robots).

\textit{Goal Space Conditioned Learning}:
To avoid dependence on reference data, the goal space methods transform the skill learning problem into a multi-objective tracking problem, thereby achieving the learning of motor skills. 
Gehring et al.\cite{gehring_hierarchical_2021} proposed a hierarchical skill learning framework, which created corresponding goal feature spaces for two different bipedal robots and manually provided target ranges. The skills learned in this way can control certain parts of the robot's body to move to specific positions in a short period. The high-level control policy uses these skills to enable robots to solve various complex tasks.
Shi et al. \cite{ETG} proposed a novel reinforcement learning based approach that consists of a foot trajectory generator. This generator continuously optimizes given tasks to provide different motion priors to guide policy learning. To handle tasks with dynamic obstacles, Bing et al. \cite{HGG} used environmental image observations to select hindsight goals and proposed a bounding-box-based hindsight goal generation (Bbox HGG). The disadvantage of these methods is that the goal space needs to be manually designed, requiring extensive expert knowledge and repeated experiments.

\textit{Unsupervised Reinforcement Learning}:
Unsupervised reinforcement learning allows robots to autonomously gather experience and learn skills using intrinsic rewards without external guidance. Some approaches, like Bellemare et al.\cite{DBLP:conf/icml/HazanKSS19}, use count-based exploration for state diversity, but this can lead to local minima. Hazan et al.\cite{bellemare2016unifying} and Liu et al.\cite{Liu2021BehaviorFT} explored maximum entropy approaches for broader state coverage, but these methods may struggle in complex Markov processes.
Other methods use mutual information to encourage diverse skill learning. Gregor et al.\cite{DBLP:journals/corr/GregorRW16} and Achiam et al.\cite{achiam2017variational} used variational inference to discover diverse behaviors. Eysenbach et al.\cite{eysenbach_diversity_2018} presented DIAYN, which uses a maximum-entropy strategy to improve exploration. However, in high-dimensional spaces, these methods can lead to static skill generation.
To address this, methods like LSD\cite{park_lipschitz-constrained_2022} and DADS\cite{DBLP:conf/iclr/SharmaGLKH20} incorporate prior knowledge to enhance dynamic skill learning. However, they often perform well only in navigation tasks. Liu et al.\cite{pmlr-v139-liu21b} integrated maximum entropy and mutual information to improve general performance across tasks like Atari games.

Unlike previous approaches, our method is inspired by the cerebellum and basal ganglia mechanisms within the CMS. By combining a hybrid reward structure with a skill activity function and discrete skill encoding, our method enables robots to learn dynamic skills that are effective across diverse downstream tasks.

\section{Preliminary}
\label{sec3}
\subsection{Reinforcement Learning} 
Reinforcement learning is an optimization method based on Markov Decision Processes (MDP) $\left\langle \mathcal{S}, \mathcal{A}, p, r\right\rangle$, where $\mathcal{S}$ represents the state space, $\mathcal{A}$ represents the action space, $p$ is the probability of  transition from one state to another after executing an action, and $r$ represents the reward. At each time step $t$, the environment provides a state  $\bm{s}_t$ to a policy $\pi$, then the policy generates an action  $\bm{a}_t \sim \pi(\bm{a}_t|\bm{s}_t)$ based on the state\cite{sutton_reinforcement_2018}. The environment executes $\bm{a}_t$ and generates the next state  $\bm{s}_{t+1} \sim p(\bm{s}_{t+1}|\bm{s}_t, \bm{a}_t)$  and a reward  $r_t=r(\bm{s}_t, \bm{a}_t, \bm{s}_{t+1})$.  From the state  $\bm{s}_t$ at time $t$ until the termination state, the sum of all discounted rewards is called the return $G_t$, and the function of the return is given by the following equation, where $\gamma$ is the discount factor between 0 and 1.
\begin{equation}
	G_t = \sum_{k=0}^{\infty}\gamma^k r_{t+k}. \label{eq1}
\end{equation}

The value $V_\pi(\bm{s}_t)$ of state $\bm{s}_t$ is defined as the mathematic expectation $\mathbb{E}_{\pi}[G_t|S_t=\bm{s}_t]$ of the return $G_t$ from that state to the termination state. Reinforcement learning seeks the optimal policy by maximizing the return expectation starting from the initial state $\bm{s}_0$. The time horizon of the reinforcement learning process in this article is finite. Within a certain time step $T$, the policy will generate a finite trajectory  $\tau=\langle\bm{s}_0,\mathbf{a}_0,\bm{s}_1,r_0,\ldots\bm{s}_{T-1},\bm{a}_{T-1},\bm{s}_T,r_{T-1}\rangle$, the goal of RL is to find the optimal policy $\pi^*$ to maximize the following objective function
\begin{align}
	{J(\pi)=\mathbb{E}_{\tau \sim {\pi}}\left[\sum_{t=0}^{T-1}{\gamma^tr_t}\right]}. \label{eq2}
\end{align}

\subsection{Unsupervised Pre-Training RL}
In unsupervised pre-training RL, the agent trains in MDP $\left\langle \mathcal{S}, \mathcal{A}, p, \mathcal{Z}\right\rangle$ without external rewards, where $\mathcal{Z}$ represents the skill space. Given a skill $\bm{z}$ and a policy $\pi(\bm{a}_t|\bm{s}_t, \bm{z})$,  the trajectory $\tau=\langle\bm{s}_0,\mathbf{a}_0,\bm{s}_1,\ldots\bm{s}_{T-1},\bm{a}_{T-1},\bm{s}_T\rangle$ can be obtained during the $T$-step interaction between the agent and the environment. The goal of this method is to enable agents to quickly adapt to downstream environments by maximizing intrinsic rewards $r^{\bm{z}}$ during the process. 

Using mutual information (MI) as $r^{\bm{z}}$ is a promising approach \cite{Liu2021BehaviorFT}\cite{eysenbach_diversity_2018}. The core idea of this method is to maximize the mutual information between the state $\bm{s}$ generated by policy $\pi$ and skill $\bm{z}$, thereby encouraging agents to correspond different skills to different states. MI rewards can be expressed as follows
\begin{align}
	I&(\bm{s};\bm{z}|\pi)=I(\bm{z};\bm{s}|\pi) \nonumber \\
	&=-\mathbb{E}_{p(\bm{z})}\left[\log p(\bm{z})\right]+ \mathbb{E}_{p(\bm{z})}\mathbb{E}_{p(\bm{s}|\pi,\bm{z})}\left[\log p(\bm{z}|\bm{s},\pi)\right] .
	\label{eq3}
\end{align}

Calculating $p(\bm{z}|\bm{s},\pi)$  during training is a tricky task. We can introduce $q(\bm{z}|\bm{s})$ as a variational approximation for this value, where $q$ can be updated by the posterior data $z$ generated during the training process \cite{barber2004algorithm}. From $\mathcal{H}(p)\le\mathcal{H}(p,q)$, the following equation can be obtained
\begin{align}
	I&(\bm{s};\bm{z}|\pi) \nonumber \\
	&\ge -\mathbb{E}_{p(\bm{z})}\left[\log p(\bm{z})\right]+ \mathbb{E}_{p(\bm{z})}\mathbb{E}_{p(\bm{s}|\pi,\bm{z})}\left[\log q(\bm{z}|\bm{s})\right].
	\label{eq4}
\end{align}

Therefore, the intrinsic reward $r^{\bm{z}}$ can be recorded as 
\begin{align}
	\label{eq5}
	r^{\bm{z}}_{t} = \log p(\bm{z}) + \log q(\bm{z}|\bm{s}_{t+1}) ,
\end{align}
then use the collected trajectory to find the optimal policy $\pi^*$ by maximizing the following objective:
\begin{align}
	\label{eq6}
	{J(\pi)=\mathbb{E}_{\tau \sim {\pi}}\left[\sum_{t=0}^{T-1}{\gamma^tr^{\bm{z}}_{t}}\right]}. 
\end{align}

\section{Methodology}
\label{sec4}
This section mainly consists of two parts. The first part introduces how to design CMS-PRL based on neurological mechanisms. The second part provides specific descriptions of each component of CMS-PRL, as well as how to pre-train the motion primitive module and use learned skills to solve different downstream tasks.

\subsection{From CMS to CMS-PRL}
\begin{figure*}[!t]
	\centering
	\includegraphics[width=7.0in]{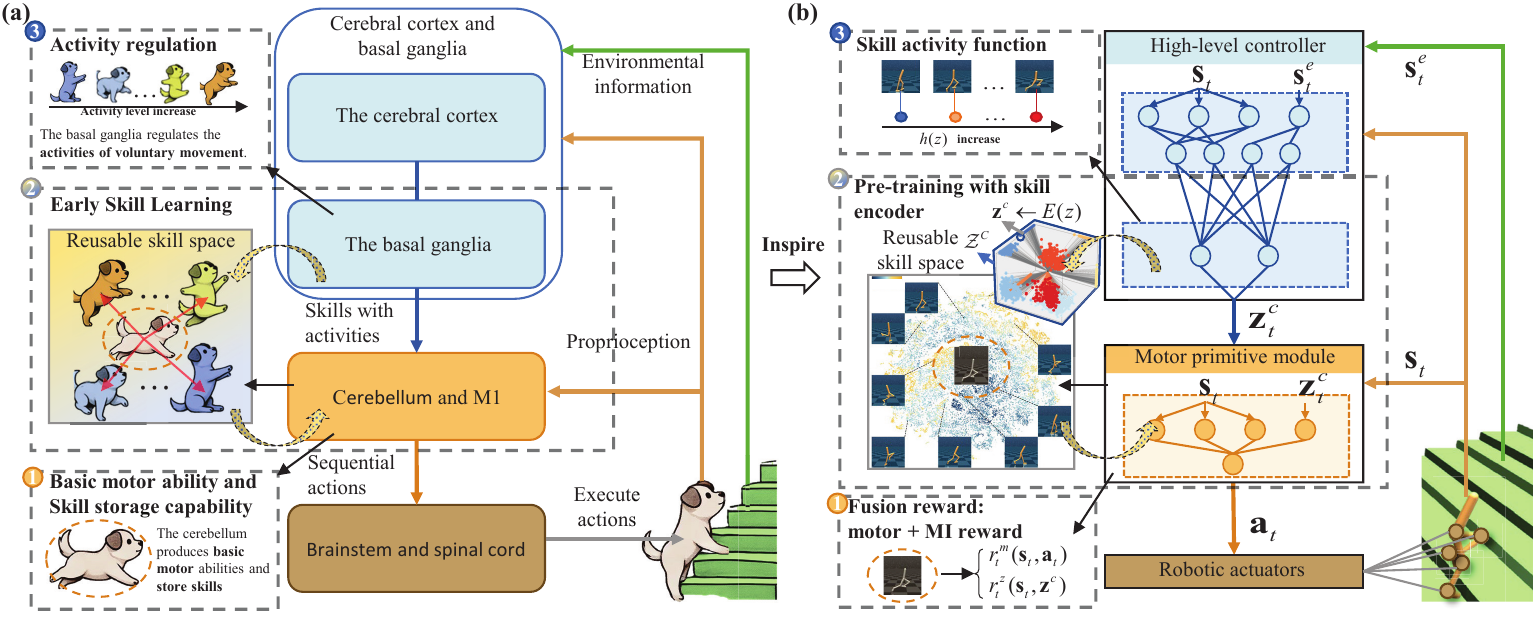}
	\caption{Overview of the proposed framework architecture. Panel (a) on the left illustrates the structure of the Central Motor System (CMS), utilizing three colors to represent its three hierarchical control centers. The three gray dashed boxes, labeled sequentially, correspond to the functions of the cerebellum, the early-stage skill learning roles of both the cerebellum and basal ganglia, and the basal ganglia's regulation of voluntary movement. Panel (b) on the right presents the architecture of the proposed CMS-PRL framework, aligned with the three hierarchical levels depicted in panel (a). The dashed boxes indicate components inspired by the cerebellum (motor primitive model's fusion reward function), skill learning (pre-training), and the basal ganglia's regulation of movement (skill activity function).}
	\label{FIG1}
\end{figure*}
This section outlines the design of the CMS-PRL algorithm, inspired by the CMS \cite{bib2}. 
As shown in Fig. \ref{FIG1}a, the CMS is structured in three layers: the high-level (cerebral cortex and basal ganglia), mid-level (cerebellum and primary motor cortex), and low-level (brainstem and spinal cord). The high-level components process environmental and proprioceptive information and provide motor programs. The cerebellum refines these programs and sends them to the primary motor cortex, while the brainstem and spinal cord execute muscle control.
The cerebellum and basal ganglia play key roles in motor learning and control:
\begin{enumerate}
	\item Cerebellar motor regulation and skill storage: The cerebellum coordinates movements and stores motor skills. It regulates movement through an internal gait-encoding mechanism \cite{powell2015synaptic}, and studies have shown its role in restoring gait and balance in stroke patients \cite{koch2019effect}. Additional research \cite{ebner2008cerebellum, bursztyn2006neural} confirms its role as an internal model for motor skill storage.
	\item Basal ganglia and cerebellum in early skill learning: The basal ganglia provide motor programs to the cerebellum during early skill learning. The subthalamic nucleus projects to the cerebellar cortex, while the dentate nucleus sends projections to the striatum \cite{bostan2018basal}. This collaboration enables motor program refinement through continued practice, with the cerebellum storing the skills \cite{senatore2012role, todorov2019interplay}.
	\item Basal ganglia regulation of voluntary movement: The basal ganglia regulate movement through direct pathways that facilitate movement and indirect pathways that inhibit it \cite{delong1990primate, albin1989functional}, enabling humans to generate flexible motor patterns. Dysfunction in these pathways can cause motor disorders, such as hypokinesia or hyperkinesia \cite{vonsattel_huntington_1998, poewe_parkinson_2017}.
\end{enumerate}

Building on these principles, we design the CMS-PRL framework (Fig. \ref{FIG1}b), which includes a high-level controller, mid-level motor primitive module, and low-level actuators. The high-level controller provides skill instructions, the mid-level stores and generates precise actions, and the low-level actuators execute them. Key mechanisms in the CMS-PRL framework are as following:
\begin{enumerate}
	\item Fusion reward: Inspired by the cerebellum's role in generating basic movements and storing skills, we design a fusion reward $r^{f}$. It consists of two parts: the basic motor reward  $r^{m}$ and the mutual information reward  $r^{z}$. $r^{m}$ facilitates the development of fundamental motor abilities in robots, such as walking and balance.  During robotic movement, $r^{z}$ aids in the discovery of diverse dynamic skills. The continuous movement of the robots ensures significant differences between states, thereby preventing the formation of static skills. These skills are stored in a continuous skill space $\mathcal{Z}^c$ for reusing by the high-level controller.
	\item Skill encoder: Mimicking the basal ganglia's role in generating motor programs, we develop a discrete skill encoder that maps discrete skills $z$ to a continuous skill space $\mathcal{Z}^c$, providing diverse skill inputs $\bm{z}^c$ to the motor primitive module during pre-training.
	\item Skill activity function: Based on the basal ganglia's regulation of movement activity, we introduce a skill activity function  $h(z)$, which calculates the activity level of different skills and uses it as a weight for the motor reward, associating different skills with specific movement patterns.
\end{enumerate}

\begin{figure}[!t]
	\centering
	\includegraphics[width=3.4in]{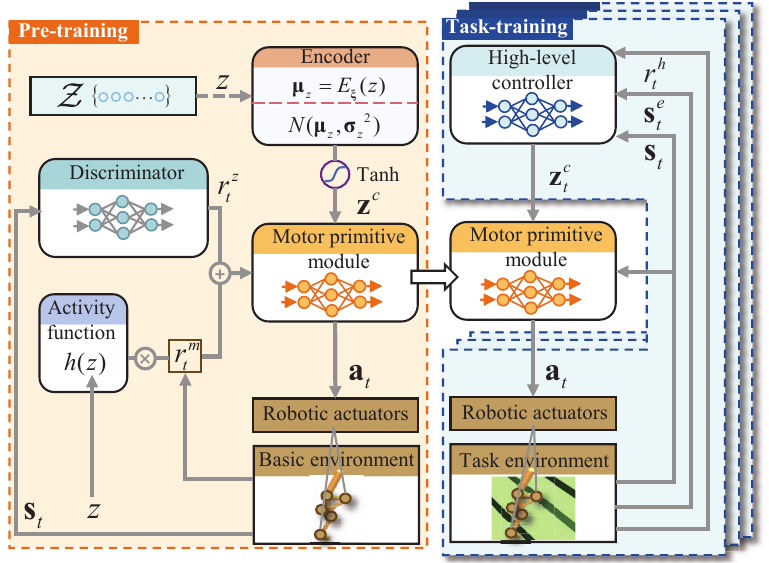}
	\caption{Learning process of the CMS-PRL algorithm. The learning process consists of two phases: In the pre-training phase, a motor primitive module is developed to receive signals from the high-level controller and adjust the robot's movement intentions and posture. During the task training phase, a high-level controller is trained to process external environmental information and generate task-specific skill signals.}
	\label{FIG2}
\end{figure}

As shown in Fig. \ref{FIG2}, the CMS-PRL framework has two training stages: pre-training and task training. The pre-training stage is conducted in the basic environments, which only include the robots themselves. During pre-training, the motor primitive module learns and stores diverse motor skills using the fusion reward, skill encoder, and skill activity function. Discrete skills $z$ are sampled from the random skill space $\mathcal{Z}$, encoded into a latent space of dimension $d_{z}$, and input to the motor primitive module.
The module outputs actions $\bm{a}_t$ based on the proprioception state $\bm{s}_t$, with the basic motor reward $r^m_t$ calculated through interaction with the environment. The skill activity function $h(z)$ weighs the motor reward, while a discriminator computes the mutual information reward $r^z_t$. The module is trained by maximizing the fusion reward $r^{f}_{t}$, as outlined in Algorithm \ref{alg1}.

The task-training stage is conducted in task environments, which consist of robots and task goals. During the task training stage, the high-level controller outputs the skill  $\bm{z}^c_t$ based on proprioception $\bm{s}_t$ and environmental information $\bm{s}^e_t$, calling the motor primitive module to execute actions and complete complex tasks. The high-level controller is trained by maximizing the sparse reward $r^h_t$, with the process detailed in Algorithm \ref{alg2}.

\subsection{The Skill Encoder}
The discrete skill space $\mathcal{Z}$ consists of $C$ different skills, $\mathcal{Z}=\{z_0,z_1,...,z_i,...z_{C-1}\}$. We use a uniform distribution $p(z)=U(0,C-1) $ to sample a skill from $\mathcal{Z}$. 

Through the one-hot encoding format, the discrete skill $z$ is first converted into a $C$-dimensional vector $\bm{v} = \delta(z)$, where $\delta$ is the kronecker delta function.
To map arbitrary one-hot skill $\bm{v}$ to the latent space $\mathcal{Z}^c$ of the $d_z$ dimension, we use a parameterized encoder $E_{\bm{\xi}}$. The latent skill $\bm{z}^c$ is generated by the following equations
\begin{align}
	\label{eq7}
	\bm{\mu} = &E_{\bm{\xi}}(\bm{v}) =  E_{\bm{\xi}}(\delta(z)),\\
	\bm{z}^c  &\sim\mathcal{N}(\bm{\mu}, \bm{\sigma}^2), 
\end{align}
where $\mathcal{N}$ is a normal distribution, $\bm{\sigma}$ is a fixed standard deviation vector. By this way, the encoded skills will be centered around $\bm{\mu}$ and dispersed into skill clusters.
To ensure the sampling efficiency, a bounded latent space is used, where $\bm{z}^c$ can be limited to $[-1, 1]$ through the processing of the $\tanh$ function.

To increase the differences between different latent skills, we update $E_{\bm{\xi}}$ by maximizing the L2-norm between continuous skill vectors, which is called the skill diversity (SD) objective, as shown in the following equation
\begin{align}
	\label{eq8}
	\max_{\bm{\xi}}L(\bm{\xi})=\sum_{i=0}^{C-2} \sum_{j=i+1}^{C-1} \|{\hat{\bm{\mu}}}_{i} - \hat{{\bm{\mu}}}_{j}\|_2,
\end{align}
where ${\hat{\bm{\mu}}}=\frac{\tanh(\bm{\mu})}{\|\bm{\mu}\|_2}$.

\subsection{The Fusion reward}
The fusion reward $r^f_t$ is shown in the following formula
\begin{align}
	r^{f}_{t}(\bm{s}_t, \bm{a}_t, \bm{z}^c) = &\beta h(z)r^{m}_{t}(\bm{s}_t,\bm{a}_t)  +(1-\beta) r^{z}_{t}(\bm{s}_t, \bm{z}^c),
	\label{eq15}
\end{align}
where $\beta \in (0,1)$ is the proportion of the basic motor reward and the mutual information reward.

In order to achieve the basic motor generation function of the cerebellum, we design a basic motor reward $r^{m}_{t}$ as following
\begin{align}
	r^{m}_{t}(\bm{s}_t,\bm{a}_t) = w_v \bm{v}_x  -w_c \|\bm{a}_t\|^2 + w_fI[\bm{s}_t \notin \bar{\bm{s}_t}] ,
	\label{eq12}
\end{align}
where $w_v$ is the the weight of torso velocity, $\bm{v}_x$ represents the robot's velocity in the x-axis, $w_f$ is the weight of balance ability, $w_c$ is the weight of energy consumption, $I[\cdot]$ is the indicator function, $\bar{\bm{s}}$ is the invalid state after a fall.  This reward promotes the robot's ability to move and maintain balance, while reducing energy consumption. At the same time, the reward value can  be used as a quantitative indicator to measure the motor activity level.

In order to achieve the skill learning and storage function of the cerebellum, we design a mutual information reward $r^{z}_{t}$  as following
\begin{align}
	r^{z}_{t}(\bm{s}_t, \bm{z}^c) = -\log p(z) + \log q_{\bm{\varphi}}(\phi(\bm{s}_{t})) ,
	\label{eq13}
\end{align}
where $q_{\bm{\varphi}}: \mathcal{S}\to \mathcal{Z}$ is a parameterized variational estimation discriminator, and $\phi$ is a feature extraction function, $z$ is the discrete skill. 
The variational estimation parameters $\bm{\varphi}$ can be updated by the following equation
\begin{align}
	\min_{\bm{\varphi}} L(\bm{\varphi}) = -\frac{1}{B} \sum^{B}_{j=1}\sum^{C-1}_{i=0} \bm{v}_{j_i} \log q_{\bm{\varphi}}(\phi(\bm{s}_j)),
	\label{eq14}
\end{align}
where ${\bm{v}}_{j}$ is the one-hot vector of $z_j$.

\subsection{The Skill Activity Function}
\begin{figure}[h]
	\centering
	\includegraphics[width=3.4in]{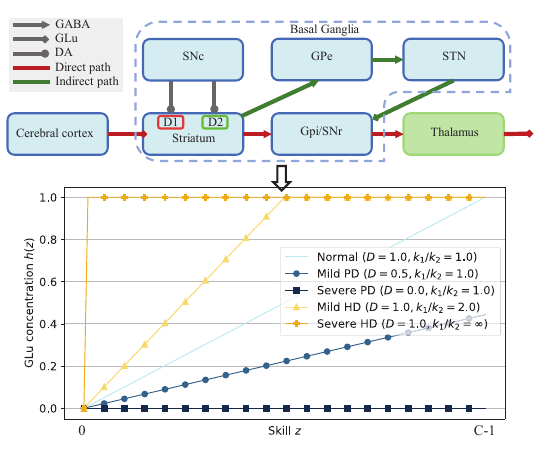}
	\caption{Schematic of the basal ganglia's internal structure and the skill activation function. Different neurotransmitters within the basal ganglia are represented by three distinct line patterns, with the direct and indirect pathways indicated in red and green, respectively. The skill activation function is depicted as a curve that varies with discrete skills, where different colors and line styles represent the activation function under both healthy and abnormal basal ganglia parameters.}
	\label{FIG3}
\end{figure}

The Basal ganglia regulates motor activity primarily through two pathways: the direct pathway, which promotes movement, and the indirect pathway, which inhibits movement\cite{delong1990primate},\cite{albin1989functional} (as shown in Fig.\ref{FIG3}). Information from the cerebral cortex is first processed by the striatum. This information is then transmitted through both the direct (striatum $\to$ globus pallidus internus (GPi)/substantia nigra pars reticulata (SNr) $\to$ thalamus) and indirect (striatum $\to$ globus pallidus externus (GPe) $\to$ subthalamic nucleus (STN) $\to$ thalamus) pathways, ultimately affecting the excitatory output of the thalamus to the motor cortex.

In the direct pathway, inhibitory signals from the neurotransmitter gamma-aminobutyric acid (GABA) reduce activity in the GPi, allowing the thalamus to release more glutamate (Glu), an excitatory neurotransmitter, to stimulate the motor cortex and facilitate movement. Conversely, the indirect pathway increases inhibition on the thalamus, reducing cortical excitation and thus inhibiting movement. Dopamine (DA), released by the substantia nigra pars compacta (SNc), modulates these pathways: it enhances the direct pathway via D1 receptors and inhibits the indirect pathway via D2 receptors, overall promoting cortical activity and facilitating movement.

When basal ganglia function is impaired, it can lead to movement disorders such as Huntington's disease (HD) or Parkinson's disease (PD)\cite{vonsattel_huntington_1998},\cite{poewe_parkinson_2017}. In HD, the indirect pathway is blocked, leading to excessive movement, while in PD, reduced dopamine release leads to impaired movement initiation.

Based on the internal antagonistic effect of the basal ganglia, we design a skill activity calculation function $h(z)$
\begin{align}
	h(z) &= min\left[H(\exp(D) \frac{k_1}{k_2} z - \exp(-D)\frac{k_1}{k_2} z),1\right], \nonumber \\
	&=min\left[H\frac{k_1 (\exp(2D) - 1)}{k_2 \exp(D)} z,1\right] ,
	\label{eq11}
\end{align} 
where $D\in[0,1]$ represents dopamine concentration, $k_1\in(0,1]$ and $k_2\in(0,1]$ respectively represent the degree of activation of direct and indirect pathways, $H$ is the normalization constant, $H = exp(1)/[(C-1)(exp(2)-1)]$. Under normal conditions, $D=1$ and $k_1/k_2 = 1$, leading to a uniform motor activity range. In PD, reduced dopamine ($D<1$) flattens the function curve, while in HD, the imbalance of the pathways ($k_1/k_2 > 1$) results in heightened activity, as shown in Fig.\ref{FIG3}.

The activity values calculated based on $h(z)$ are used as the weights of the motor reward $r^m_t$. Under the influence of weights, the motor primitive module will learn skills corresponding to different motor rewards, thus exhibiting different levels of activity.

\subsection{The Pre-training stage}
In the pre-training stage, the mid-level motor primitive module is modeled as a reinforcement learning policy $\pi$, with the optimization objective shown in the following equation
\begin{align}
	\max_{\pi} \mathbb{E}_{p(\bm{z}^c)}\mathbb{E}_{p(\tau|\pi_{},\bm{z}^c)}\left[\sum_{t=0}^{T-1}\gamma^tr^{f}_{t}(\bm{s}_t, \bm{a}_t, \bm{z}^c)\right].
	\label{eq16}
\end{align}

We use the Soft Actor-Critic algorithm (SAC)\cite{haarnoja_soft_2018} to implement the iterative process of optimization. SAC improves the exploration capability of the policy by introducing a state entropy term into the optimization objective. Therefore, the optimization objective becomes
\begin{align}
	\max_{\pi} \mathbb{E}_{p(\bm{z}^c)}\mathbb{E}_{p(\tau|\pi_{},\bm{z}^c)}\left[\sum_{t=0}^{T-1}\gamma^tr^{f}_{t}(\bm{s}_t, \bm{a}_t, \bm{z}^c) + \alpha H(\pi(\cdot|\bm{s}_t, \bm{z}^c))\right] ,
	\label{eq16}
\end{align}
where $\alpha$ is a regularization coefficient to control the importance of entropy.

We use the Soft-Q value to calculate the value of $\pi$ at each iteration, as shown in the following equation
\begin{align}
	Q(\bm{s}_t,\bm{a}_t,\bm{z}^c) = r^{f}_{t}(\bm{s}_t, \bm{a}_t, \bm{z}^c) + \gamma V(\bm{s}_{t+1}) ,
	\label{eq17}
\end{align}
where 
\begin{align}
	V(\bm{s}_{t}) = \mathbb{E}_{\bm{a}_t \sim \pi(\cdot|\bm{s}_t,\bm{z}^c)}[Q(\bm{s}_t, \bm{a}_t, \bm{z}^c) - \alpha \log \pi(\bm{a}_t|\bm{s}_t, \bm{z}^c)].
	\label{eq18}
\end{align}

In order to achieve policy iteration in continuous space, we use two neural networks as approximations for the $\pi$ and $Q$, $\pi_{\bm{\theta}}(\bm{a}_t|\bm{s}_t,\bm{z}^c)$ and  $Q_{\bm{\rho}}(\bm{s}_t,\bm{a}_t,\bm{z}^c)$, respectively. Then we can learn $Q_{\bm{\rho}}$ by using the following equation
\begin{align}
	&\min_{\bm{\rho}} J(\bm{\rho}) =
	{\mathbb{E}}_{\substack{(\bm{s}_{t},\bm{a}_{t}, \bm{z}^c)\sim \mathcal{B}}} 
	\left[\frac{1}{2}\left( 
	Q_{\bm{\rho}}(\bm{s}_t, \bm{a}_t,\bm{z}^c) - G \right)^2\right],
	\label{eq19}
\end{align}
where $G =  r^{f}_{t}(\bm{s}_t, \bm{a}_t, \bm{z}^c) + \gamma V(\bm{s}_{t+1}) $. While $\pi_{\bm{\theta}}$ can be learned through the following equation
\begin{align}
	&\min_{\bm{\theta}}  J(\bm{\theta})  =   \nonumber \\
	&\mathbb{E}_{(\bm{s}_t,\bm{z}^c) \sim \mathcal{B}}
	\left[
	{\mathbb{E}_{\bm{a}_t\sim {\pi _{\bm{\theta}}}}}
	\left[
	\alpha \log \pi _{\bm{\theta}}(\bm{a}_t|\bm{s}_t, \bm{z}^c) - {Q_{\bm{\rho}}}(\bm{s}_t ,\bm{a}_t, \bm{z}^c)
	\right]
	\right].
	\label{eq20}
\end{align}

In summary, Algorithm \ref{alg1} provides the pseudo code for the pre-training process. 
\begin{algorithm}
	\captionsetup{justification=centering} 
	\caption{Pre-training pseudo code}\label{alg1}
	\begin{algorithmic}[1]
		\STATE \textbf{Input:} maximum number of episodes $N_1$;  maximum number of time-step $T$; discount factor $\gamma$; proportion of reward $\beta$; batch size $B$; discrete skills number $C$; encoder standard deviation $\sigma$; number of update episodes $n$; learning rates $\eta_1$ for ${\bm{\xi}}$ $\bm{\varphi}$, $\bm{\theta}$, $\bm{\rho}$.
		
		\STATE \textbf{Output:} optimal parameters $\bm{\theta}^*$ for the mid-level policy;
		\STATE \textbf{Init:} $\pi_{\bm{\theta}}\gets$ mid-level policy; $E_{\bm{\xi}}\gets$ Encoder; $q_{\bm{\varphi}}\gets$ discriminator; $Q_{\bm{\rho}}\gets$ Q-function; $\mathcal{B}\gets$ Buffer;
		\FOR{$i\gets 1,N_1$}
		\STATE  Sample skill $z\sim p(z)= U(0, C-1)$ 
		\STATE  Get one-hot vector $\bm{v} =\delta(z)$ 
		\STATE  $done$ = False
		\FOR{time-step $t=0,...,T-1$}
		\STATE Get latent skill $\bm{z}^c\sim N(E_{\bm{\xi}}(\bm{v}), \bm{\sigma}^2)$
		\STATE $\bm{z}^c\gets \tanh (\bm{z}^c)$
		\STATE Sample action  $\bm{a}_t \sim \pi_{\bm{\theta}}(\bm{a_t}|\bm{s}_t,\bm{z}^c)$
		\STATE ${\bm{s}_{t + 1}},r^{m}_{t},done\sim p({\bm{s}_{t + 1}}|{\bm{s}_t},{\bm{a}_t})$
		\STATE Compute $r^{f}_{t}$ according to Equation (\ref{eq15})
		\STATE Store $({\bm{s}_t},{\bm{s}_{t + 1}},{\bm{a}_t},\bm{z}^c, z, r^{f}_{t})$  in  $\mathcal{B}$
		\STATE ${\bm{s}_t} \gets {\bm{s}_{t+1}}$
		\IF{$done$}
		\STATE $break$
		\ENDIF
		\ENDFOR
		\FOR{update-step $u= 1,...,n$}  
		\STATE Get all discrete skills $(z_0,...,z_{C-1})$
		\STATE Compute unit vectors $(\hat{\bm{\mu}} _{0},...,\hat {\bm{\mu}}_{{C - 1}})$
		\STATE Update  ${\bm{\xi}}$ by Equation (\ref{eq8})
		\STATE Sample B transitions $b_B$ from $\mathcal{B}$ 
		\STATE Update  ${\bm{\varphi}}$ by Equation (\ref{eq14})
		\STATE Update  ${\bm{\rho}}$ by Equation (\ref{eq19})
		\STATE Update  ${\bm{\theta}}$ by Equation (\ref{eq20})
		
		\ENDFOR
		\ENDFOR
	\end{algorithmic}
\end{algorithm}

\subsection{The Task-training Stage}
In the task-training stage, we model the high-level controller as a reinforcement learning policy $\psi$.
After obtaining the motor primitive module policy  $\pi_{\bm{\theta}^*}(\bm{a}|\bm{s},\bm{z}^c)$, the low-dimensional continuous skill space $\mathcal{Z}^c$ can be used to learn the high-level policy $\psi(\bm{z}^c|\bm{s},\bm{s}^e)$. In order to maintain the consistency of the reward in different tasks, sparse rewards $r^{h}_{t}$ are used in the task environment, and robots only receive positive rewards when completing a specified goal. Therefore, in this environment, robots need to have strong exploration ability to complete tasks.

According to the characteristics of the CMS, the high-level policy has strict temporal abstraction, which means switching skill actions $\bm{z}^{c}_{t}$ every $k$ time steps, while the motor primitive module policy generates different basic actions $\bm{a}_t$ each time step. This temporal abstraction can lead to a decrease in the training efficiency of the high-level policy, as only $T/k$ transitions can be obtained within the $ T $ time step. Therefore, we adopt a step conditioned critical structure as described in  \cite{gehring_hierarchical_2021},\cite{whitney_dynamics-aware_2019}. By using this architecture, we can obtain $T-(k-1)$ transformed data within $ T $ time steps to increase the number of samples and ensure training efficiency under the constraints of time abstraction.

We still use SAC as the iterative optimization algorithm, due to the use of step-conditioned critic, some improvements need to be made to the Q-value function $Q(\bm{s},\bm{s}^e, \bm{z}^c)$ and its optimization objectives. To simplify calculations, $\bm{s}_t,\bm{s}^e_t$ are denoted as $\bm{s}^+_t$ together. The improved Q-value function adds input $i=t\mod k(0\le i<k)$ to record the number of executed steps for $\bm{z}^c_t$. The improved Q-value function is shown as follow
\begin{align}
	Q(\bm{s}_t^ + ,\bm{z}_t^c,i) =
	\left(\sum_{j = 0}^{k - i - 1} \gamma ^j r^{h}_{t + j}\right) + \gamma ^{k - i} V \left(\bm{s}_{t + k - i} ^{+} \right),
	\label{eq21}
\end{align}
where 
\begin{align}
	V \left(\bm{s}_{t}^ +\right)= \mathbb{E}_{\bm{z}_t^c\sim \psi} \left[Q(\bm{s}_t^ + ,\bm{z}_t^c,0) - \alpha \log \psi (\bm{z}_t^c|\bm{s}_t^ + )\right] .
	\label{eq22}
\end{align}

Similarly, we use parameterized soft Q-value function $Q_{\bm{\upsilon}}(\bm{s}_t^ +,\bm{z}_t^c, i)$ and policy $\psi_{\bm{\omega}}(\bm{z}^c|\bm{s}_t^ +)$, with neural network parameters of $\bm{\upsilon}$ and $\bm{\omega}$, respectively. The parameters $\bm{\upsilon}$ of Q-function can be trained using the following equation
\begin{align}
	&\min_{\bm{\upsilon}} J(\bm{\upsilon}) =\nonumber \\
	&{\mathbb{E}}_{\substack{(\bm{s}_{t,...,t + k - 1}^ + ,\bm{z}_t^c,i)\sim \mathcal{B}}} 
	\left[\frac{1}{2}\left( 
	Q_{\bm{\upsilon}}(\bm{s}_t^ + ,\bm{z}_t^c,i) - G^{'} \right)^2\right],
	\label{eq23}
\end{align}
where
\begin{align}
	G^{'} = 
	\sum_{j = 0}^{k - i - 1} (\gamma ^j r^{h}_{t + j}) 
	+ \gamma ^{k - i} V (\bm{s}_{t + k - i} ^{+} ).
	\label{eq2344}
\end{align}
And the optimization target for the high-level policy parameters $\bm{\omega}$ is shown in the following equation
\begin{align}
	&\min_{\bm{\omega}}  J(\bm{\omega}) =  \nonumber \\
	&\mathbb{E}_{\bm{s}_t^ + \sim \mathcal{B}}
	\left[
	{\mathbb{E}_{\bm{z}_t^c\sim {\psi _{\bm{\omega}}}}}
	\left[
	\alpha \log \psi _{\bm{\omega}}(\bm{z}_t^c|\bm{s}_t^ + ) - {Q_{\bm{\upsilon}}}(\bm{s}_t^ + ,\bm{z}_t^c,0)
	\right]
	\right] .
	\label{eq24}
\end{align}

Algorithm \ref{alg2} provides the pseudo code for the task-training process. 

\begin{algorithm}
	\captionsetup{justification=centering} 
	\caption{Task-training pseudo code}\label{alg2}
	\begin{algorithmic}[1]
		\STATE \textbf{Input:} 
		maximum number of episodes $N_2$;  maximum number of time-step $T$; discount factor $\gamma$; batch size $B$; number of update episodes $n$; high action interval $k$; learning rates $\eta_2$ for $\bm{\upsilon}$, $\bm{\omega}$.		
		\STATE \textbf{Output:} optimal parameters  $\bm{\omega}^*$ for high-level policy;
		\STATE \textbf{Init:} $\pi_{\bm{\theta^{*}}}\gets$ learned mid-level policy;  $Q_{\bm{\upsilon}}\gets$ Q-function;
		$\psi_{\bm{\omega}}\gets$ high-level policy; $\mathcal{B}\gets$ Buffer;
		
		\FOR{$i\gets 1,N_3$}
		\STATE  $done$ = False
		\FOR{time-step $t=0,...,T-1$}
		\STATE Compute $i = (t\mod{k})$
		
		\IF{$i==0$}
		\STATE Sample high-action  $\bm{z}_t^c\sim \psi _{\bm{\omega}}(\bm{z}_t^c|{\bm{s}_t},\bm{s}_t^e)$
		\ENDIF
		
		\STATE Sample action  $\bm{a}_t\sim {\pi _{\bm{\theta}^*}}({\bm{a}_t}|{\bm{s}_t},\bm{z}_t^c)$
		\STATE ${\bm{s}_{t + 1}},\bm{s}_{t + 1}^e,r^{h}_{t},done\sim p({\bm{s}_{t + 1}},\bm{s}_{t + 1}^e|{\bm{s}_t},\bm{s}_t^e,{\bm{a}_t})$
		\STATE Store  $({\bm{s}_t},{\bm{s}_{t + 1}},\bm{s}_t^e,\bm{s}_{t + 1}^e,\bm{z}_t^c,{r^{h}_{t}},i)$ in $\mathcal{B}$
		\STATE ${\bm{s}_t},\bm{s}_t^e \leftarrow {\bm{s}_{t + 1}},\bm{s}_{t + 1}^e$
		
		\IF{$done$}
		\STATE $break$
		\ENDIF
		\ENDFOR
		\FOR{update-step $u= 1,...,n$}  
		\STATE Sample B transitions $b_B$ from $\mathcal{B}$ 
		\STATE Update ${\bm{\upsilon}}$ using Adam according to Equation (\ref{eq23})
		\STATE Update ${\bm{\omega}}$ using Adam according to Equation (\ref{eq24})
		\ENDFOR
		
		\ENDFOR
	\end{algorithmic}
\end{algorithm}

\section{Experiments}
\label{sec5}
\subsection{Tasks and Baselines}
\begin{figure}[!t]
	\centering
	\includegraphics[width=3.2in]{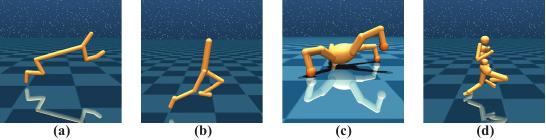}
	\caption{Different types of robots in simulation environments. 
		(a) Cheetah. (b) Walker. (c) Quadruped. (d) Humanoid.}
	\label{FIG4}
\end{figure}

\begin{figure}[!t]
	\centering
	\includegraphics[width=3.2in]{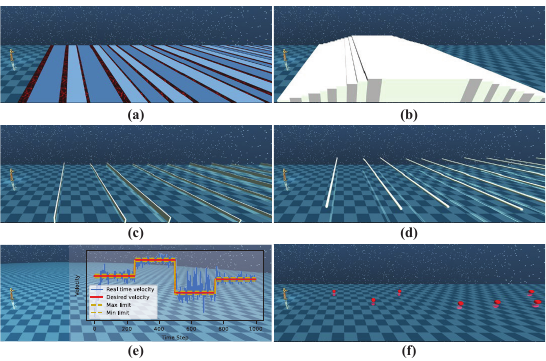}
	\caption{Different task environments. (a) Gaps: The robot needs to constantly move forward and cross the magma, but once the robot comes into contact with the magma or falls, the task ends. (b) Stairs: The robot needs to cross the steps of ascent and descent. (c) Hurdles: The robot needs to adjust its posture to cross the hurdles. (d) Limbos: The robot needs to lower its center of gravity during movement to pass through the interceptor. (e) V-track: The robot needs to track four random expected speeds and maintain its own speed within a specified deviation range from the expected speed. (f) Goals: The robot needs to touch more target points within a limited time.}
	\label{FIG5}
\end{figure}
This section presents a series of experiments to evaluate the CMS-PRL algorithm against popular baselines across various robots and tasks.
We build the robot motion simulation environments using the Mujoco physics engine \cite{2012MuJoCo}. Four robots from dm\_control\cite{tunyasuvunakool2020dm_control} are used as control subjects, and two additional tasks include target tracking and velocity tracking are added based on the multi-class sparse-reward tasks from Bisk\cite{hsd3_github}.

1) Robot types: As shown in Fig. \ref{FIG4}, the robots include Cheetah, Walker, Quadruped, and Humanoid. Cheetah and Quadruped move in the XZ plane, while Walker and Humanoid operate in three-dimensional space. Control difficulty increases from Cheetah to Humanoid.
Following the classic Gym \cite{Gym} setup, each robot's state includes joint positions, velocities, and normalized contact forces, while actions are represented by joint torques. The state-action dimensions are as follows: Cheetah (59, 6), Walker (59, 6), Quadruped (145, 12), and Humanoid (149, 21). 

For Quadruped and Humanoid, we preprocess their states with a feature function $\phi$\cite{gehring_hierarchical_2021} to help the discriminator better capture body part coordination when calculating the mutual information reward $r^z_t$. 
For Quadruped, the swing decomposition method \cite{dobrowolski2015swing} is used to compute torso rotation angles and toe distances relative to the torso. For Humanoid, we calculate body part positions relative to the pelvis. No specific feature extraction is applied to Cheetah and Walker.

2) Task types: As shown in Fig. \ref{FIG5}, tasks such as Gaps, Stairs, Hurdles, and Limbos follow the Bisk settings \cite{hsd3_github}. Additionally, two new tasks, V-track and Goals, are introduced:
\begin{itemize}
	\item V-track: The robot must adjust its actions to track randomly sampled target speeds every 250 time steps. A reward of 1 is given if the robot's speed is within $0.1 m/s$ of the target.
	\item Goals: The robot must reach as many target points as possible within a time limit. A reward of 1 is given if the robot's center of mass is within 0.05 m of the target point.
\end{itemize}

All tasks use sparse reward $r^h_t$, where a reward of $+1$ is given for completing the task, $-1$ for entering an invalid state, and $0$ otherwise.

To demonstrate the competitiveness of the proposed method in these tasks, we use three popular algorithms as baseline methods for comparison in our experiments: SAC\cite{haarnoja_soft_2018}, a stable policy-based maximum entropy reinforcement learning method; DIAYN\cite{eysenbach_diversity_2018}, a representative unsupervised reinforcement learning method that utilizes intrinsic rewards based on mutual information. In our implementation, the discrete skills are processed by a linear layer and output continuous skills with $d_z$ dimension, so this baseline is called DIAYN-C; APS\cite{pmlr-v139-liu21b}, a recent state-of-the-art unsupervised reinforcement learning method that combines the benefits of mutual information and maximum entropy.
These baselines cover both traditional high-performance, classic reinforcement learning methods and representative pre-training unsupervised reinforcement learning methods.

\subsection{Implementation Details}

During pre-training, we construct the motor primitive module using a 4-layer deep dense neural network\cite{sinha2020d2rl} with 256 neurons per layer. The encoder is a single-layer neural network.

For task training, the high-level controller is also built using a deep dense neural network. Proprioceptive and environmental information are processed separately: the proprioceptive input goes through a 4-layer neural network with 256 neurons per layer, while the environmental input passes through a 2-layer network with 64 neurons per layer.

Regarding the implementation of the baseline algorithms, SAC is based on OpenAI Spinning Up\cite{spinning_up_github}, while DIAYN-C and APS are implemented based on URBL\cite{url_benchmark_github}. All algorithms use the same network architecture and configuration as the proposed method to ensure fairness. The hyperparameters used during pre-training are shown in Table \ref{Params1}, and those used during task-training are shown in Table \ref{Params2}.
All pre-training runs with three random seeds, while task-training runs with two random seeds, resulting in six different random seed combinations. All experiments are conducted on a server equipped with an NVIDIA RTX 3090, using Python 3.7 and PyTorch 1.10.

\begin{table}
	\captionsetup{justification=centering} 
	\centering
	\caption{Hyper-parameters for pre-training}\label{Params1}
	\begin{tabularx}{\columnwidth}{@{}Xl@{}}
		\toprule
		Parameter & Value \\
		\midrule
		Episodes number $N_1$    &   $2 \cdot 10^5$  \\
		Batch size $B$    &   $256$    \\
		Horizon $T$     & 100 \\
		Update numbers $n$ & 50  \\
		Number of skills $C$  		& 10  \\
		Latent skill dim $d_{z}$	& 7  \\
		Torso velocity weight $w_v$  & 1.0 \\
		Balance ability weight $w_f$ & 0.1 \\
		Energy consumption weight $w_c$ & 0.1  \\
		Proportion of reward  $\beta$ & 0.5  \\
		Dopamine concentration $D$ & 1.0  \\
		Degree of activation of direct pathway $k_1$ & 1.0 \\
		Degree of activation of indirect pathway $k_2$ & 1.0 \\
		Std of skill coding  $\sigma_{z}$ & 0.3  \\
		Learning rates $\eta$ &  $3 \cdot 10^{-4}$   \\
		Discount factor $\gamma$ & 0.99 \\
		\bottomrule
	\end{tabularx}
\end{table}

\begin{table}
	\captionsetup{justification=centering} 
	\centering
	\caption{Hyper-parameters for task-training}\label{Params2}
	\begin{tabularx}{\columnwidth}{@{}Xl@{}}
		\toprule
		Parameter & Value \\
		\midrule
		Episodes number $N_2$    &   $1 \cdot 10^4$  \\
		Batch size $B$      &   $256$  \\
		Horizon $T$     & 1000 \\
		Update numbers $n$ & 50  \\
		Learning rates $\eta$ &  $3 \cdot 10^{-4}$   \\
		Discount factor $\gamma$ & 0.99 \\
		Action interval $k$ & 3 \\
		\bottomrule
	\end{tabularx}
\end{table}

\subsection{Learning results of downstream tasks}

\begin{table}[h]
	\captionsetup{justification=centering} 
	\centering
	\caption{comparison of our method (CMS-PRL) with baselines. Mean rewards (scores) and standard deviations over six training runs}\label{tab3}
	\begin{threeparttable}
	\resizebox{\columnwidth}{!}{
		\begin{tabular}{l c c c c c}
			\toprule
			\multicolumn{2}{c}{Environment} & \multicolumn{3}{c}{Baseline} & \multicolumn{1}{c}{Ours} \\
			\cmidrule(lr){1-2} \cmidrule(lr){3-5} \cmidrule(lr){6-6}
			Robot&Task &SAC &APS&DIAYN-C&CMS-PRL\\
			\midrule 
			\multirow{5}{*}{Cheetah}&Gaps&0.0 $\pm$ 0.0& \textbf{16.1} $\pm$ 1.7&5.5 $\pm$ 6.4 &7.9 $\pm$ 0.4\\
			&Goals& \textbf{27.5} $\pm$ 18.6&11.5 $\pm$ 3.2&3.1 $\pm$ 1.3&10.9 $\pm$ 0.5\\
			&Stairs&9.9 $\pm$ 11.4&18.8 $\pm$ 0.6&18.8 $\pm$ 0.9&\textbf{19.5} $\pm$ 0.3\\
			&Hurdles&0.0 $\pm$ 0.1&\textbf{31.7} $\pm$ 0.6  &22.4 $\pm$ 3.8 &29.3 $\pm$ 2.1\\
			&V-track& \textbf{964.6} $\pm$ 10.3&345.5 $\pm$ 145.7&352.5 $\pm$ 49.6&568.0 $\pm$ 73.0\\
			\hline
			\multirow{6}{*}{Walker}&Gaps&0.0 $\pm$ 0.0&0.0 $\pm$ 0.0&0.0 $\pm$ 0.0&\textbf{17.8} $\pm$ 2.3\\
			&Goals&0.0 $\pm$ 0.0&0.0 $\pm$ 0.0 &0.0 $\pm$ 0.0&\textbf{9.3} $\pm$ 0.5\\
			&Stairs&16.5 $\pm$ 1.9&0.1 $\pm$ 0.2&\textbf{16.9}$\pm$ 0.9&13.9 $\pm$ 1.9\\
			&Hurdles& \textbf{17.7} $\pm$ 3.3&0.0 $\pm$ 0.0&4.8 $\pm$ 4.7&15.5 $\pm$ 0.5\\
			&V-track& \textbf{877.5} $\pm$ 16.0&255.5 $\pm$ 53.4 &324.1 $\pm$ 36.5&790.3 $\pm$ 24.3\\
			&Limbos&5.3 $\pm$ 7.0&0.0 $\pm$ 0.0&0.0 $\pm$ 0.0&\textbf{9.0} $\pm$ 2.5\\
			\hline
			\multirow{5}{*}{Quadruped}&Gaps&0.0 $\pm$ 0.0&0.0 $\pm$ 0.0&0.0 $\pm$ 0.0&\textbf{4.5} $\pm$ 0.4\\
			&Goals&0.0 $\pm$ 0.0&0.0 $\pm$ 0.0&0.0 $\pm$ 0.0&\textbf{13.6} $\pm$ 0.3 \\ 
			&Stairs&0.0 $\pm$ 0.0&0.0 $\pm$ 0.0&0.0 $\pm$ 0.0&\textbf{14.3} $\pm$ 2.5\\
			&Hurdles&0.0 $\pm$ 0.0&0.7 $\pm$ 1.0&0.0 $\pm$ 0.0&\textbf{5.0} $\pm$ 3.7\\
			&V-track& \textbf{359.7} $\pm$ 98.2&53.5 $\pm$ 24.6&55.2 $\pm$ 9.0&294.3 $\pm$ 132.9\\
			\hline
			\multirow{6}{*}{Humanoid}&Gaps&0.0 $\pm$ 0.0&0.0 $\pm$ 0.0&0.0 $\pm$ 0.0& \textbf{2.2} $\pm$ 2.2\\
			&Goals&0.0 $\pm$ 0.0&0.0 $\pm$ 0.0&0.0 $\pm$ 0.0&\textbf{5.4} $\pm$ 0.9\\
			&Stairs&0.0 $\pm$ 0.0&0.0 $\pm$ 0.0&0.2 $\pm$ 0.4& \textbf{7.9} $\pm$ 1.1\\
			&Hurdles&0.0 $\pm$ 0.1&0.0 $\pm$ 0.1&0.4 $\pm$ 0.5&\textbf{2.7} $\pm$ 0.2\\
			&V-track&\textbf{455.7} $\pm$ 113.4&95.3 $\pm$  14.3&161.9 $\pm$ 19.1&342.3 $\pm$  64.6\\
			&Limbos& 0.2 $\pm$ 0.4&0.0 $\pm$ 0.0&0.0 $\pm$ 0.0&\textbf{3.0} $\pm$ 1.0\\
			\bottomrule
		\end{tabular}
	}
	
\end{threeparttable}
\end{table}



\begin{figure}[!t]
	\centering
	\includegraphics[width=3.4in]{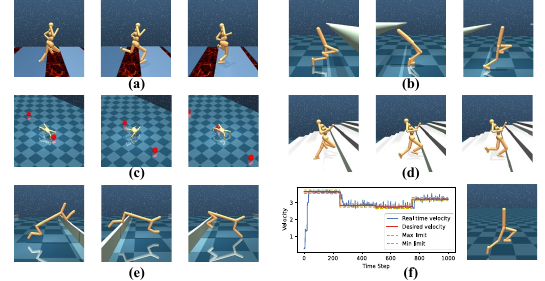}
	\caption{
		The control performance of CMSRL algorithm in different robot task environments.
		(a) Humanoid robot in the Gaps task environment. 
		(b) Walker robot in the Limbos task environment.
		(c) Quadruped robot in the Goals task environment.
		(e) Humanoid robot in the Stairs task environment.
		(f) Cheetah robot in the Hurdles task environment.
		(g) Walker robot in the V-track task environment.
	}
	\label{FIG6}
\end{figure}

This section compares the performance of CMS-PRL with different baseline algorithms in various downstream tasks.
Table \ref{tab3} presents the performance scores for different robots and tasks. CMS-PRL demonstrates superior adaptability, achieving effective scores in all 22 test environments without notable weaknesses.

As robot complexity increases, CMS-PRL shows a more pronounced advantage over baseline methods. While simpler robots with low-dimensional state spaces perform well with baseline algorithms, performance declines as complexity increases. For instance, SAC excels in the Goals and V-track tasks for Cheetah, while APS performs best in Gaps and Hurdles. For the more complex Walker robot, DIAYN-C performs well in Stairs, but CMS-PRL takes the lead in Gaps, Goals, and Limbos, while SAC continues to do well in V-track and Hurdles.
 
With three-dimensional robots like Quadruped and Humanoid, only SAC performs well in V-track, while CMS-PRL achieves high scores across all other tasks. SAC outperforms in velocity tracking tasks across all robots, indicating that skills learned through unsupervised methods may have limitations in precise control. However, CMS-PRL consistently ranks second, highlighting its strength in motor control compared to other pre-training methods.

Fig. \ref{FIG6} illustrates the performance of CMS-PRL across various robots and downstream tasks, highlighting the advanced motor capabilities of each robot:
\begin{itemize}
	\item Fig. \ref{FIG6}a:  Humanoid robot demonstrates effective jumping behavior in the Gaps task, leaping over lava-filled gaps.
	\item Fig. \ref{FIG6}b:  Walker robot adjusts posture to overcome obstacles taller than itself.
	\item Fig. \ref{FIG6}c:  Quadruped robot skillfully navigates multiple targets in the Goals task, accurately tracking moving and stationary objects.
	\item Fig. \ref{FIG6}d:   Humanoid robot adeptly climbs steep stairs, adjusting its leg and body coordination.
	\item Fig. \ref{FIG6}e: Cheetah robot efficiently leaps over hurdles, maintaining high-speed movement.
	\item Fig. \ref{FIG6}f: Walker robot demonstrates consistent velocity regulation over time.
\end{itemize}

These results highlight CMS-PRL's effectiveness in developing advanced motor skills, enabling robots to manage complex tasks with precision and flexibility.

\subsection{Ablation experiments}
To clarify the impact of key internal components in the proposed method, we conduct ablation experiments:
\begin{enumerate}
	\item \textit{No $h(z)r^m$}: This variant eliminates the motor reward weighted by the skill activity function, retaining only the mutual information reward while maintaining the remainder of the method unchanged.
	\item \textit{No SD}: This variant excludes the skill diversity objective from the discrete skill encoding process, with the rest of the method remaining intact.
\end{enumerate}

Table \ref{tab4} presents the results of the ablation experiments. Comparing the performance of \textit{No $h(z)r^m$} with CMS-PRL reveals a significant performance drop across most tasks when the weighted motor reward is removed. For instance, in Walker’s Gaps task, the score decreases from 17.8 to 0, highlighting the critical role of the motor reward in helping the robot develop the fundamental motor skills needed to adapt to complex and dynamic environments. Without this component, the robot lacks the necessary coordination and control.

On the other hand,  \textit{No SD} performs relatively well in several tasks, especially with simpler robots like Cheetah and Walker. For example, in Cheetah's Gaps task,  \textit{No SD} achieves a score of 12.5, surpassing CMS-PRL's score of 7.9. Similarly, in the Goals task for the Walker, \textit{No SD} attains a score of 10.1, compared to 9.3 for the proposed algorithm. However, \textit{No SD} struggles in more complex tasks. For instance, it fails to perform effectively in Humanoid's Gaps task, where CMS-PRL completes the task successfully.

These results suggest that for robots with low-dimensional state spaces,the skill diversity objective may lead to unnecessary exploration, consequently diminishing the algorithm's performance. Conversely, for high-dimensional robots like the Humanoid, incorporating the skill diversity objective appears essential, as it fosters broader exploration and aids in addressing more challenging tasks.

\begin{table}[h]
	\captionsetup{justification=centering} 
	\centering
	\caption{comparison of our method (CMS-PRL) with ablation methods. Mean rewards (scores) and standard deviations over six training runs}\label{tab4}
	\begin{threeparttable}
		\resizebox{\columnwidth}{!}{
			\begin{tabular}{l c c c c}
			\toprule
			\multicolumn{2}{c}{Environment} & \multicolumn{2}{c}{Ablation} & \multicolumn{1}{c}{Ours} \\
			\cmidrule(lr){1-2} \cmidrule(lr){3-4} \cmidrule(lr){5-5}
			Robot&Task &No SD&No $h(z)r^m$ &CMS-PRL\\
			\midrule 
			\multirow{5}{*}{Cheetah}&Gaps&\textbf{12.5} $\pm$ 1.7& 0.7 $\pm$ 1.4&7.9 $\pm$ 0.4\\
			&Goals& \textbf{13.8} $\pm$ 1.6&3.9 $\pm$ 2.6&10.9 $\pm$ 0.5\\
			&Stairs&\textbf{19.7} $\pm$ 0.1&19.0 $\pm$ 1.0&19.5 $\pm$ 0.3\\
			&Hurdles&\textbf{32.0} $\pm$ 3.0&16.4$\pm$ 1.9   &29.3 $\pm$ 2.1\\
			&V-track& \textbf{718.9}$\pm$ 68.5&276.2 $\pm$ 32.7&568.0 $\pm$ 73.0\\
			\hline
			\multirow{6}{*}{Walker}&Gaps&\textbf{19.6} $\pm$ 3.5&0.0 $\pm$ 0.0&17.8 $\pm$ 2.3\\
			&Goals&\textbf{10.1} $\pm$ 1.5&0.0 $\pm$ 0.0 &9.3 $\pm$ 0.5\\
			&Stairs&\textbf{17.0} $\pm$ 0.8&0.3 $\pm$ 0.3&13.9 $\pm$ 1.9\\
			&Hurdles& \textbf{15.6} $\pm$ 2.3&0.6 $\pm$ 0.7&15.5 $\pm$ 0.5\\
			&V-track& \textbf{796.4} $\pm$ 6.5&145.6 $\pm$ 196.5&790.3 $\pm$ 24.3\\
			&Limbos&5.3 $\pm$ 2.7&0.0 $\pm$ 0.0&\textbf{9.0} $\pm$ 2.5\\
			\hline
			\multirow{5}{*}{Quadruped}&Gaps&\textbf{6.7} $\pm$ 0.8&0.0 $\pm$ 0.0&4.5 $\pm$ 0.4\\
			&Goals&\textbf{17.1} $\pm$ 0.4&0.0 $\pm$ 0.0&13.6 $\pm$ 0.3 \\ 
			&Stairs&\textbf{16.7} $\pm$ 0.6&0.0 $\pm$ 0.0&14.3 $\pm$ 2.5\\
			&Hurdles&\textbf{7.5} $\pm$ 0.7&0.0 $\pm$ 0.0&5.0 $\pm$ 3.7\\
			&V-track& 214.7$\pm$ 48.8&79.1 $\pm$ 32.1&\textbf{294.3} $\pm$ 132.9\\
			\hline
			\multirow{6}{*}{Humanoid\tnote{$\bullet$}}&Gaps\tnote{$\ast$}&0.3 $\pm$ 0.5&0.0 $\pm$ 0.0& \textbf{2.2} $\pm$ 2.2\\
			&Goals&\textbf{6.1} $\pm$ 1.7&0.0 $\pm$ 0.0&5.4 $\pm$ 0.9\\
			&Stairs&5.8 $\pm$ 3.2&0.2 $\pm$ 0.4& \textbf{7.9} $\pm$ 1.1\\
			&Hurdles&\textbf{3.0} $\pm$ 1.4&0.4 $\pm$ 0.5&2.7 $\pm$ 0.2\\
			&V-track&\textbf{487.3} $\pm$ 55.4&169.1 $\pm$  91.2&342.3 $\pm$  64.6\\
			&Limbos& 1.8 $\pm$ 0.9&0.0 $\pm$ 0.0&\textbf{3.0} $\pm$ 1.0\\

			\bottomrule
		\end{tabular}
	}
	\begin{tablenotes}
		\scriptsize
		\item[$\bullet$] The robot with the highest complexity among the four types of robots.  
		\item[$\ast$] The most difficult task among all types of tasks, as stated in reference\cite{gehring2021hierarchical},
		
		has no algorithm that can effectively score. 
	\end{tablenotes}
\end{threeparttable}
\end{table}

\subsection{Skills learned by the motor primitive module}
\begin{figure}[!t]
	\centering
	\includegraphics[width=3.2in]{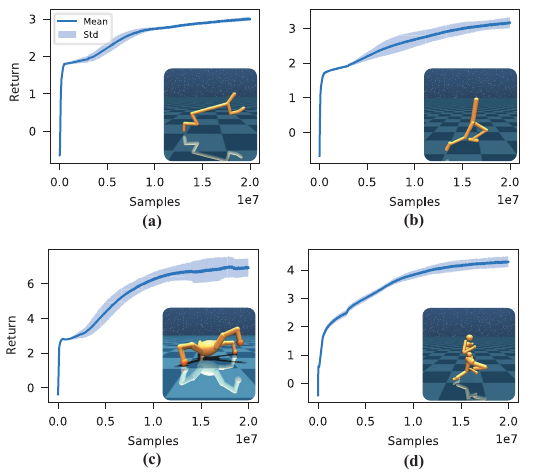}
	\caption{
		The $r^{f}$ return images of the four robots over time during the pre-training process. The curve represents the average value obtained from three sets of random experiments, and the shaded portion represents the standard deviation.
		(a). Cheetah. (b). Walker. (c). Quadruped. (d). Humanoid.}
	\label{FIG7}
\end{figure}

\begin{figure}[!t]
	\centering
	\includegraphics[width=3.3in]{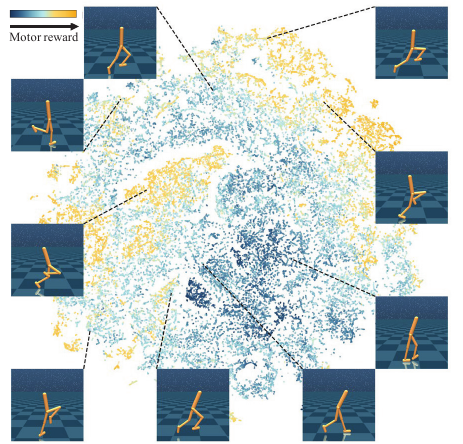}
	\caption{The state-skill pairs t-SNE scatter plot obtained by the motor primitive module of the Walker robot under the action of random skills from the $\mathcal{Z}^c$. }
	\label{FIG8}
\end{figure}
The results of both the comparative and ablation experiments demonstrate that CMS-PRL enables different types of robots to adapt effectively to various downstream tasks using their learned skills. This adaptability is attributed to the diverse levels of active motor skills learned by the motor primitive module during pre-training. In this section, we evaluate and analyze the effectiveness of the motor primitive module in learning these skills during the pre-training stage.

Pre-training continues until the return of the fusion reward $r^{f}$ converges. Fig. \ref{FIG7} presents the mean and standard deviation of fusion rewards for different robots across iterations. These statistics were calculated over three learning sessions, each comprising 20 million samples. The results indicate that all four robots successfully achieved reward convergence during pre-training.

To further evaluate the motor primitive module, we conduct a random skill input experiment in basic environments. We randomly sample 200 skills $\bm{z}^c$ from the skill space $\mathcal{Z}^c$ as inputs to the motor primitive module. Under the influence of these skills, the robots interact with environments and receive corresponding basic motor rewards $r^m$. Using $r^m$ as a measure of motor activity, we apply the t-SNE dimensionality reduction method \cite{van2008visualizing} to project the hidden vectors obtained by inputting $(\bm{s}_t, \bm{z}^c)$ into the policy $\pi_{\bm{\theta}}$ into a two-dimensional space.

Fig. \ref{FIG8} presents the t-SNE scatter plot for the Walker robot, where each point represents a state-skill pair $(\bm{s}_t, \bm{z}^c)$, and the color of each point corresponds to the motor reward $r^m$. The images adjacent to the points depict the robot's posture associated with the respective skill. The figure clearly demonstrates that different random skills induce a uniform spectrum of motor activities, ranging from low to high, with each skill corresponding to a distinct motor pattern. This observation confirms that the motor primitive module effectively learned a diverse array of dynamic skills during the pre-training stage.

\subsection{Adjustment effect of the skill activity function}
\label{secEPC}

Based on the motor regulation mechanisms of the basal ganglia, we propose a skill activity calculation function, as illustrated in Equation \ref{eq11}. By adjusting the dopamine concentration parameter $D$ and the activation level parameters $k_1$ and $k_2$ for the two pathways, we can modulate the output GLu concentration, thereby influencing the motor activity of the robot. 

Preliminary experiments show that under normal parameter settings, the skill encoding method can use skills to regulate the various motor intentions of the motor primitive module, enabling it to respond flexibly to various skills.  To further explore the impact of the skill activity function under abnormal basal ganglia parameter conditions, we designed the following experiment. 

\begin{figure}[h]
	\centering
	\includegraphics[width=3.3in]{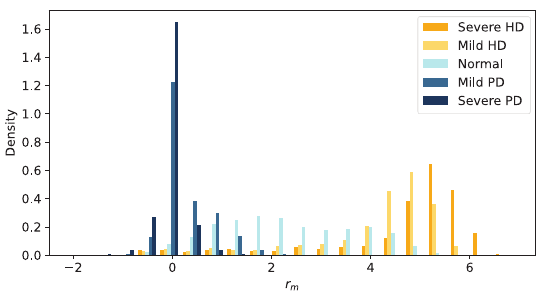}
	\caption{The distribution of basic motor rewards $r^m$ generated by the motor primitive modules of the Walker robot under different skill activity calculation functions.}
	\label{FIG9}
	
\end{figure}

Following the abnormal scenarios depicted in Fig. \ref{FIG4}, we develop multiple parameter sets, as detailed in Table \ref{tab5}. By employing these variations of the skill activity calculation function during the pre-training stage, we obtain corresponding motor primitive modules. We then randomly select 200 skills  $\bm{z}^c$  from the skill space $\mathcal{Z}^c$  as inputs for the policies. For each skill, the Walker robot interacts with the environment for 250 steps, during which we record multiple motor rewards $r^m$. The distribution of these rewards is illustrated in Fig. \ref{FIG9}.

In Huntington’s disease (HD)-like conditions, characterized by high overall skill activity, motor rewards $r^m$ are predominantly distributed toward the higher end, resulting in the robot exhibiting uncontrollable movements. Conversely, in Parkinson's disease (PD)-like conditions, where dopamine deficiency leads to reduced overall skill activity, $r^m$ values cluster toward the lower end, causing the robot to struggle with initiating movement effectively. 

These findings indicate that abnormal configurations of the skill activity calculation function can produce motor behaviors in robots that mimic symptoms of basal ganglia-related disorders, such as HD and PD. This suggests that the skill activity function is crucial for modulating motor control and underscores its importance in understanding and simulating motor dysfunction in robotic systems.

\begin{table}[h]
	\captionsetup{justification=centering} 
	\centering
	\caption{Parameter Table for Abnormal Skill Activity Calculation Function}\label{tab5}%
	\begin{tabularx}{\columnwidth}{@{}lccccc@{}}
		\toprule
		Parameter & Sever HD & Mild HD & Normal & Mild PD & Sever PD \\
		\midrule
		$D$   & 1.0   & 1.0   & 1.0   & 0.5  & 0.0 \\
		$k_1$ & 1.0   & 1.0   & 1.0   & 1.0  & 1.0 \\
		$k_2$ & 0.0   & 0.5   & 1.0   & 1.0  & 1.0 \\
		\bottomrule
	\end{tabularx}
\end{table}

\section{Conclusion}
\label{sec7}
In this study, we propose a pre-training reinforcement learning method (CMS-PRL) inspired by the mechanisms of the Central Motor System (CMS), aimed at enabling various types of robots to learn motor skills and adapt to a range of downstream tasks. The core concept of CMS-PRL is to emulate the functions of the cerebellum and basal ganglia within the CMS, abstracting these functions into a fusion reward, a skill encoding module, and a skill activity function. These components, integrated with reinforcement learning algorithms, enhance the robots' ability to acquire motor skills. CMS-PRL offers several key advantages: without relying on external data or expert knowledge, robots can learn a rich set of dynamic skills during pre-training through the internal fusion reward. Additionally, the motor reward, weighted by skill activity, increases the robot's adaptability to various downstream tasks, thereby improving its performance in complex environments. We evaluate the proposed method using four robots of varying complexity and six different downstream tasks within the Mujoco simulation environment. Experimental results demonstrate that CMS-PRL outperforms traditional reinforcement learning and unsupervised skill learning algorithms across a wide range of tasks.

Despite its clear advantages, CMS-PRL has some limitations that warrant further investigation. First, in precision tasks such as velocity tracking, our method—like other unsupervised skill learning approaches—exhibits lower accuracy. Moreover, the inherent complexity of neural system mechanisms implies that relying solely on reinforcement learning may overlook critical details of these biological processes. Addressing these two shortcomings represents an important direction for future research.

In future work, we propose that the incorporation of motor primitive models derived from spinal cord circuits, such as Central Pattern Generators (CPGs)\cite{bellegarda2022cpg,10611128}, could effectively replace the low-level controllers currently utilized in our algorithm. By integrating more detailed biological insights, we aim to enhance the performance of our method and broaden its applicability to real-world robotic systems.

\bibliographystyle{IEEEtranmod}
\bibliography{bibs}

\vspace{11pt}

\vfill

\end{document}